\documentclass[letterpaper, 10 pt, conference]{ieeeconf} 

\IEEEoverridecommandlockouts

\usepackage{graphicx}
\usepackage{textgreek}
\usepackage{amsmath}
\usepackage{caption}
\usepackage{subcaption}
\usepackage{booktabs}
\usepackage{multirow}
\usepackage{array}
\usepackage{hyperref}
\usepackage{algpseudocode}
\usepackage{algorithm}

\title{\LARGE \bf
SwarmHawk: Self-Sustaining Multi-Agent System for Landing on a Moving Platform through an Agent Supervision 
}

\author{Ayush Gupta, Ekaterina Dorzhieva, Ahmed Baza, Mert Alper, Aleksey Fedoseev, and Dzmitry Tsetserukou
 \thanks{*The reported study was funded by RFBR and CNRS according to the research project No. 21-58-15006.}
\thanks{The authors are with the Intelligent Space Robotics Laboratory, Space CREI, Skolkovo Institute of Science and Technology, Moscow, Russian Federation.
 {\tt \{ayush.gupta, ekaterina.dorzhieva, ahmed.baza, mert.alper, aleksey.fedoseev, d.tsetserukou\}@skoltech.ru}}
}

\begin{document}

\maketitle
\thispagestyle{empty}
\pagestyle{empty}


\begin{abstract}

Heterogeneous teams of mobile robots and UAVs are offering a substantial benefit in an autonomous exploration of the environment. Nevertheless, although joint exploration scenarios for such systems are widely discussed, they are still suffering from low adaptability to changes in external conditions and faults of swarm agents during the UAV docking. We propose a novel vision-based drone swarm docking system for robust landing on a moving platform when one of the agents lost its position signal. 

The proposed SwarmHawk system relies on vision-based detection for the mobile platform tracking and navigation of its agents. Each drone of the swarm carries an RGB camera and AprilTag3 QR-code marker on board. SwarmHawk can switch between two modes of operation, acting as a homogeneous swarm in case of global UAV localization or assigning leader drones to navigate its neighbors in case of a camera fault in one of the drones or global localization failure.

Two experiments were performed to evaluate SwarmHawk's performance under the global and local localization with static and moving platforms. The experimental results revealed a sufficient accuracy in the swarm landing task on a static mobile platform (error of 4.2 cm in homogeneous formation and 1.9 cm in leader-follower formation) and on moving platform (error of 6.9 cm in homogeneous formation and 4.7 cm in leader-follower formation). Moreover, the drones showed a good landing on a platform moving along a complex trajectory (average error of 19.4 cm) in leader-follower formation.
 
The proposed SwarmHawk technology can be potentially applied in various swarm scenarios, including complex environment exploration, inspection, and drone delivery.

\end{abstract}

\section{Introduction}

\begin{figure}[!h]
 \includegraphics[width=1.0\linewidth]{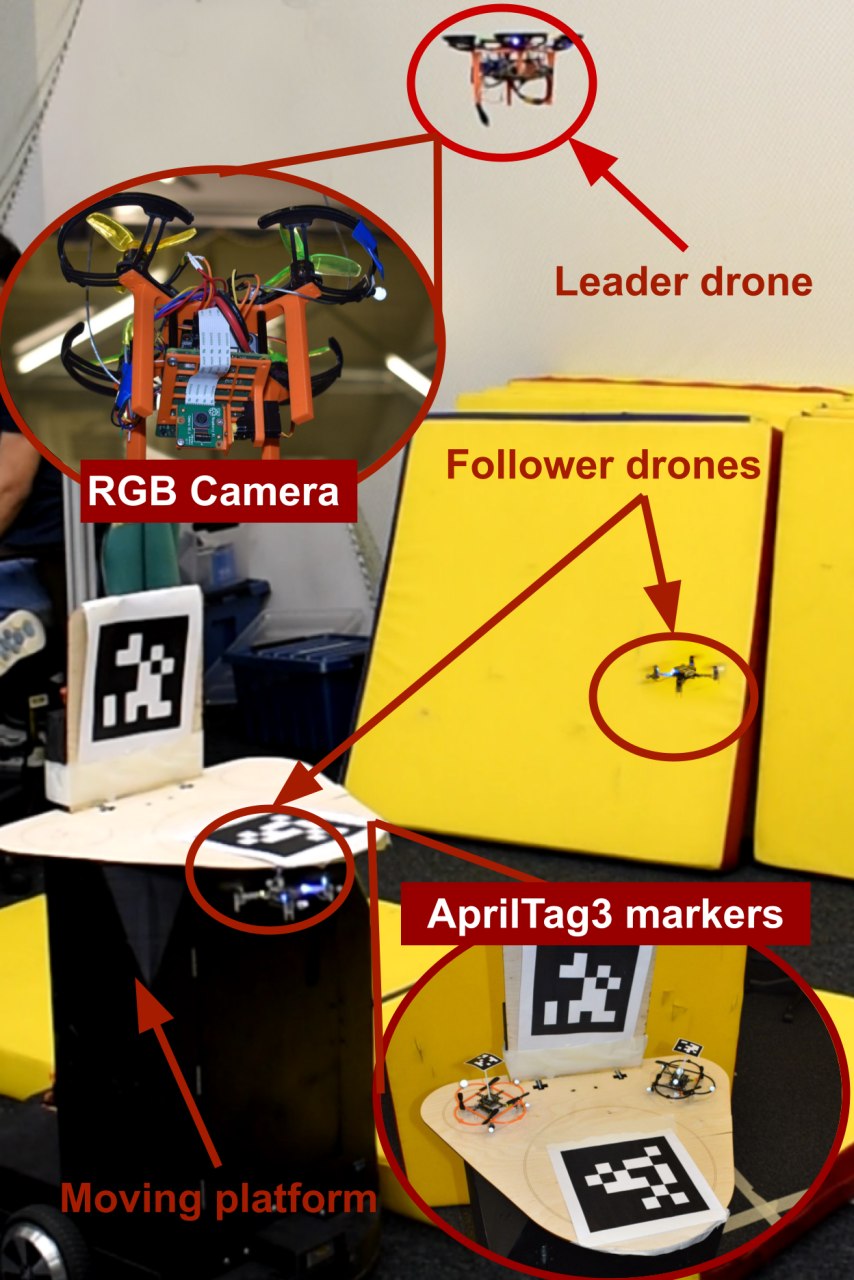}
 \caption{SwarmHawk allows landing of the swarm on moving platform under the leader-drone supervised localization.}
 \label{fig:main}
\end{figure}




The multi-agent system has been a broad research area for the last decade with the application of these systems in search and rescue, sensor networks in the Internet of Things (IoT), remote sensing surveillance, mapping, inspection, and etc. \cite{Kangunde_2021}. The autonomous delivery solutions with heterogeneous mobile robot teams are extensively investigated by delivery companies such as Amazon Technologies Inc., presenting the concepts of multi-robotic ground delivery \cite{Scott_2020} and fulfillment center with UAV delivery \cite{Curlander_2017}. The drone swarms in the above-presented application potentially can work autonomously and save costs and time in hazardous or complex environments.

The landing of drones on the moving object requires a fast and accurate detection of the site area. Several scenarios of UAV landing with high tolerance to the surface were taken into the condition in the prior research. Sarkisov et al. \cite{Sarkisov_2018} proposed the multi-copter landing gear with robotic legs to allow drones to adapt to the surface level. The torque sensors embedded in the legs will enable this system to maintain stable landing based on the evaluation of contact force with the ground. Kornatowski et al. \cite{Kornatowski_2017} suggested an origami-inspired cargo drone with a foldable cage to achieve safety while interacting with humans and landing. Although the aforementioned approaches require the assistance of a human operator during the deployment and docking of the swarm, they also sum up the addition of more payload for the drone. Alternative methods of the landing of drones were proposed. Rothe et al. \cite{Rothe_2019} suggested the midair landing in which the drone was captured with a net carried by cooperative UAVs. Midair docking of the swarm with a robot arm equipped with a force-sensitive soft gripper is explored by Fedoseev et al. \cite{Fedoseev_2021}. Tanaka et al. \cite{Tanaka_2019} introduced a UAV parcel handover system with a high-speed vision system for supply station control and midair catching or docking of drone.

The computer vision (CV) system also plays a significant role in swarm landing scenarios for the real-time detection of the landing targets. Visually distinctive tags \cite{Falanga_2017} and color-based detection \cite{Respall_2019} algorithms were applied for the recognition of the moving platform. Other works combine the CV and GPS algorithms. For example, Feng et al. \cite{Feng_2018} proposed to observe the landing platform by the camera at a close distance and to rely on GPS data otherwise. The proposed systems do not consider the loss of data transmission from the on-board camera that eventually results in the landing accident. 

In this paper, we propose a multi-agent system SwarmHawk shown in Fig. \ref{fig:main} that is capable of drone swarm landing on a moving platform. If the drone in the swarm loses its visual transmission, the other drones support the local positioning and landing of the malfunctioning unit on the moving platform with the supervision of their on-board cameras.

\section{Related Works}
Due to high maneuverability and speed, drones have been utilized in wide range of applications, including environmental monitoring, entertainment, and delivery \cite{8120115}. However, safe flight requires expensive and often cumbersome sensors for drone localization and obstacle tracking. For example, Zhou et al. \cite{Zhou_2022} developed a fully autonomous swarm able to obtain a high-quality trajectory within a few milliseconds by data fusion from grayscale and depth camera, inertial measurement unit (IMU), and ultra-wideband sensor. Using the camera for state estimation and perception tasks minimizes the use of heavy equipment in industries, including manufacturing and construction \cite{doi:10.1146/annurev-control-060117-105149}.

When localizing by the camera, the external motion capture system and GPS may not be used, allowing drones to function in places unavailable for communication, such as mines or remote forest areas. State-of-the-art methods solve the problem of navigating without the global positioning system by using vision-based algorithms \cite{7515271}.

Visual odometry, used for onboard state estimation, allows to achieve decentralized swarm control. Weinstein et al. \cite{8276634} further improved this method with the simultaneous processing of information from visual sensors an IMU. In decentralized swarm architecture, drones in the swarm act independently from each other to avoid collision. Hence, the loss of the camera feed on one drone can damage the whole swarm. The concept of a leader drone to synchronize the motion of the swarm agents is proposed by Soon-Jo Chung et al. \cite{8424838} for localization of a drone in case of its sensors' malfunction. However, the delays in calculation of the position and trajectory on the onboard computer does not allow drones to build optimal formations during the flight. Montijano et al. \cite{7430346} proposed a solution to this problem with two simultaneous consensus controllers to calculate the relative orientations and the relative positions between swarm agents. 

In addition to localization, visual information is also required to avoid obstacles along the drone's path or to prevent collisions with other drones in the swarm. In the case of dynamic obstacles, visual feed is used to estimate the position and velocity of the obstacle. Thus, the drone can build a trajectory based on the estimated future position of the obstacles \cite{lin2020robust}. High accuracy in determining the coordinates of several obstacles is possible using depth images. However, a highly obstructed environment or weather conditions cause object recognition errors \cite{8575255, 8080161}. Special tags are sometimes used in the mentioned conditions, for example, to indicate the landing site \cite{8088164}.

To increase the range and accuracy of the target vehicle recognition, Xing et al. \cite{Xing_2019} proposed a composite landmark. Robust autonomous take-off and landing of the drone on a moving platform were developed by Palafox et al. \cite{Palafox_2019}. Alijani et al. \cite{Alijani_2020} implemented a mathematical approach in the X-Y plane based on the inclination angle and state of the UAV for a safe landing on a moving vehicle. Kalinov et al. \cite{Kalinov_2019} introduced a system with high-precision visual infrared (IR) marker recognition on a mobile robot with impedance-based control for the soft landing of a drone in a heterogeneous robotic system. 

The influences of external disturbances, such as vibration and turbulent wind, on the UAV and the landing platform during precision landing are partially explored, e.g., authors \cite{Xuan-Mung_2020} introduced a robust altitude control algorithm, landing target state estimator, and an autonomous precision landing planner to overcome disturbances. Landing in turbulent wind conditions was studied by Paris et al. \cite{Paris_2020}.

While the prior developed systems investigated the landing of a single drone, a swarm landing presents an additional challenge for the real-time system. The concept of a UAV swarm docking was suggested by Tahir et al. \cite{Tahir_2020}.

Thus, in this paper, we propose a novel vision-based swarm docking system for
robust landing on a moving platform when one of the agents lost its position signal. All drones and the moving platform are equipped with tags, according to which their coordinates are calculated.

\section{SwarmHawk System Overview}

The overall system consists of a heterogeneous robots, which are a swarm of drones and a mobile robot. Each agent of the swarm of drones is equipped with a Raspberry Pi 4 (4GB RAM), camera (8MP Pi camera Ver. 2.1) and a rotating mechanism allowing the camera to be used both horizontally and vertically depending on the task. For the localization of the drones, the VICON Vantage V5 motion capture system (12 IR cameras) was used. In addition, the mobile robot has a landing platform with two AprilTag markers that are tracked by the swarm of drones. The swarm can land on the platform, switching between two scenarios of the vision-based tracking, which are homogeneous-swarm visual localization and leader-follower visual localization. 

\subsection{Homogeneous-Swarm Visual Localization Scenario}

\begin{figure}[!h]
 \includegraphics[width=1.0\linewidth]{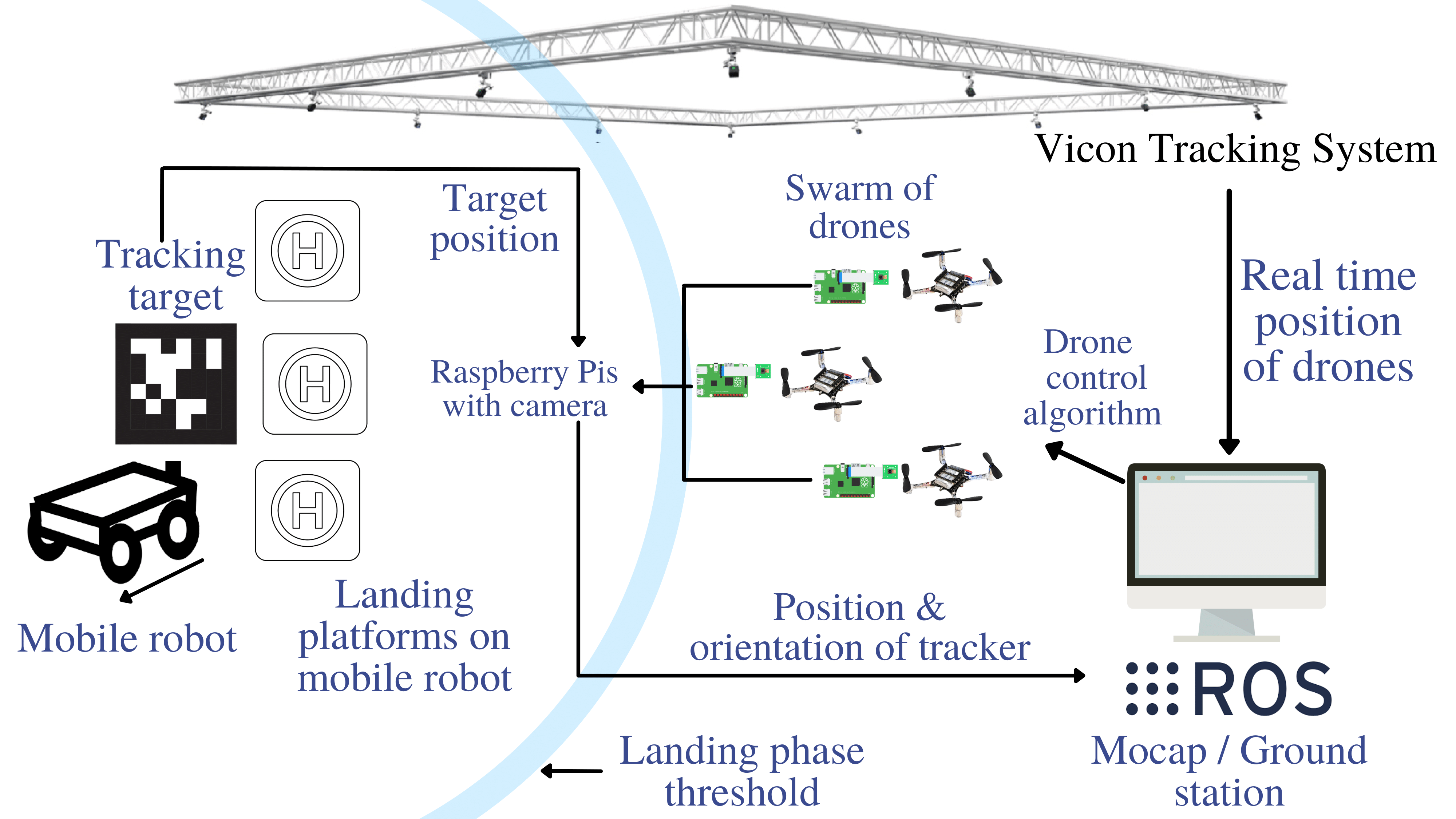}
 \caption{SwarmHawk system architecture for homogeneous-swarm visual localization scenario.}
 \label{fig:homogeneous_system}
\end{figure}

In the first scenario, i.e., homogeneous-swarm visual localization, shown in Fig. \ref{fig:homogeneous_system}, each agent of the swarm is equipped with a working Raspberry Pi camera placed in a forward-facing position.

The camera data was used by the vision-based tracking system sending the location of the AprilTag marker to the ground station with drone control algorithm operating by ROS. Therefore, each agent performs homogeneous visual localization of its own, and they move as a swarm with the help of the drone control algorithm executed on the ground station.

\subsection{Leader-Follower Visual Localization Scenario}

\begin{figure}[!h]
 \includegraphics[width=1.0\linewidth]{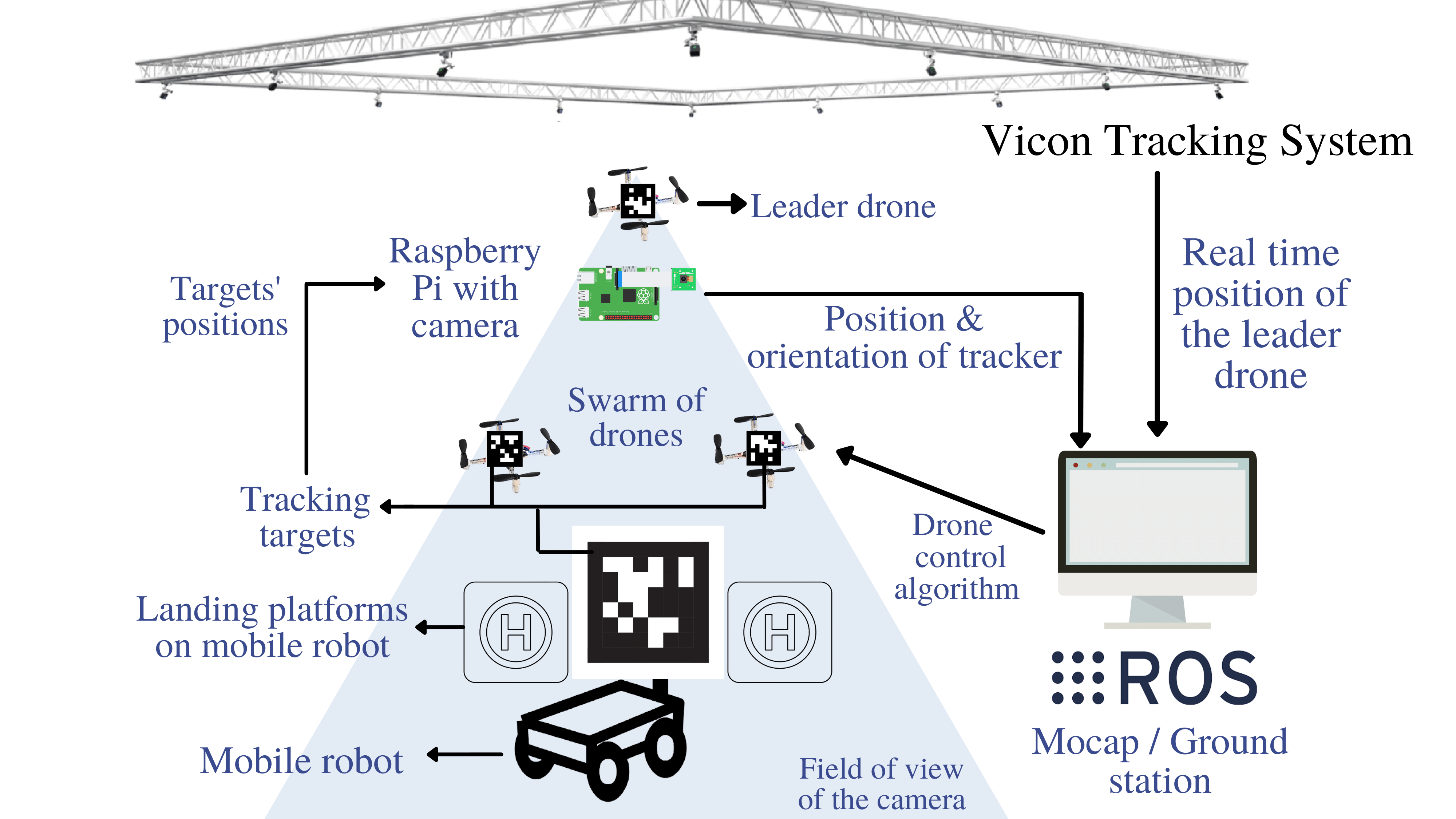}
 \caption{System architecture of leader-follower visual localization scenario.}
 \label{fig:leader-follower_system}
\end{figure}

If any of the agents lose its visual transmission, the second scenario, i.e., leader-follower visual localization, shown in Fig. \ref{fig:leader-follower_system} is initiated. 

The algorithm works such that one of the agents with a working camera becomes the leader drone, and its camera turns into a downward-facing position by a rotating mechanism. The closest ``blinded" drones then are getting assigned as followers. Since the visual sensor fault in the swarm may occur randomly, the follower drones are assigned to the leader in the formation based on the minimum value of a cost function, where Euclidean distance from the malfunctioning drone position to the leader drone position is used as a cost. The leader drone then tracks the follower drones with the help of the vision-based tracking system and the AprilTag markers placed on the drones (described in the following section). Their location are sent to the ground station handling the drone control algorithm, thus enabling the drones that lost their visual transmission to land.

\subsection{Vision-based Swarm Localization}
The goal of vision-based UAV tracking is to estimate the further trajectory of the unit relying on its position recognized in every frame of an image sequence. The correct estimation of all Cartesian positional coordinates plays a particularly important part during simultaneous landing of the swarm on a single landing pad with limited size. Therefore, the most recent visual fiducial system AprilTag3 with OpenCV library \cite{apriltag} was applied for the target detection both on the landing site of the moving platform and on the tags of the drones. ROS framework serves as a bridge to pass the coordinates of fiducial markers from onboard controllers to the ground station. We estimate the position of marker tags on the image with a graph-based image segmentation algorithm and obtain a visual estimation of possible tags more accurately than GPS systems. In addition, we tested the algorithm with tags of different sizes and opted for the size of 16.6$\times$16.6 cm based on the tag recognition performance. The detection of the tag placed on the  moving platform was confirmed at the minimum distance to the drone of 30 cm and at the maximum distance of approximately 3.5 to 4.0 m. 

\begin{figure}[!h]
 \centering
 \includegraphics[width=1.0\linewidth]{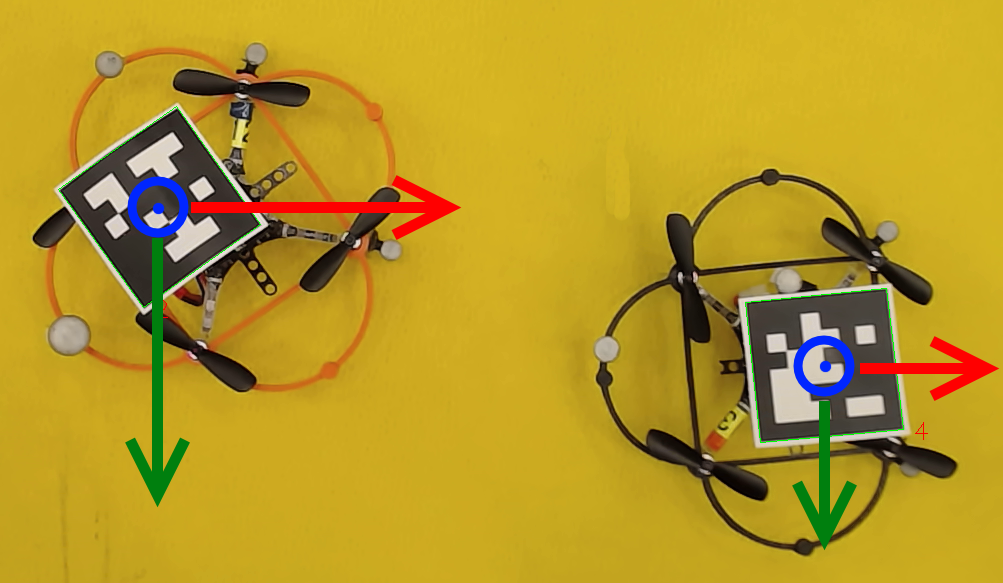} 
 \caption{AprilTag marker positions and orientations on the image frame obtained by the leader drone camera.}
 \label{fig:drones_tag}
\end{figure}

It is required to implement camera calibration as vibrations in the system can compromise the accuracy of the data. Our experiment employed digital video stabilization using the point feature matching method \cite{323794}, which works by setting motion estimation, motion smoothing, and image composition. This method is based on a 2D motion model applying a Euclidean transformation integrating translation, rotation, and scaling.

By tracking several feature points between two consecutive frames of a video, we can robustly estimate their motion and then compensate for it. The tracking algorithm works better with textured regions with corners, assuming that pixel intensities of an object are the same with every frame of the video, and we can track similar motion in neighboring pixels. To find $E(u, v)$, i.e. the difference in intensity for a pixel displacement of $(u, v)$ in all directions between the original and moved window, one of several known methods can be applied, for example, Lucas-Kanade Optical Flow method \cite{Lucas_1981}. The calculations are defined by the following equations:


\vspace{-0.4em}
\begin{equation}
 \label{eq:1}
 E(u, v) = \begin{bmatrix}
 u & v
 \end{bmatrix} M \begin{bmatrix}
 u\\
 v
 \end{bmatrix}
\end{equation}
\vspace{0.5em}

\vspace{-0.4em}
\begin{equation}
 \label{eq:2}
 M = \sum_{x, y} w(x, y) \begin{bmatrix}
 I_{x}^2 & I_{x}I_{y} \\
 I_{x}I_{y} & I_{y}^2
 \end{bmatrix},
\end{equation}
where $M$ is the $2\times2$ matrix computed from image derivatives, $u$ is the translation in $x$ direction, $v$ is the translation in $y$ direction, $w(x,y)$ is the windowing function that computes the weighted sum.


After finding good features and applying optical flow to track them, a rigid transform that maps the previous frame to the current frame of the video by using two sets of points is calculated. As soon as the system obtains a visual estimation of the motion, it can store translation and rotation (angle) in an array to obtain a smooth trajectory of motion and then smooth transforms. The latter ones can successfully stabilize the frames of the video.

\subsection{Swarm Control Algorithm}
In this work, we used a centralized swarm control approach. Thus, all path planning and decision-making are computed on the ground station while sensing and preliminary data processing is computed onboard. The swarm consisting of three drones has two modes of operation: the first is active one when all agents have working forward-facing cameras; the second one is active when some agents experiencing damages of camera or video data communication. To test these modes, we selected swarm formation flight as the primary multi-agent operation. We assumed the worst-case scenario of the mission, where all agents but one have lost the camera feed. During the first mode, each drone uses a forward-facing camera to localize the landing zone marked with AprilTag. By calculating drone path with the Artificial Potential Field (APF) introduced in \cite{APF} as a planning policy, each agent avoids collision with other drones during landing or formation flight. Each drone then is modeled as an obstacle to other units in the swarm. APF calculations are defined by Eq. (\ref{eq:3}-\ref{eq:4}):

\vspace{-0.4em}
\begin{equation}
 \label{eq:3}
 U(x,y,z) =U_{a}(x,y,z) +U_{r}(x,y,z)
\end{equation}
\vspace{0.5em}

\vspace{-0.4em}
\begin{equation}
 \label{eq:4}
U_{a}(x,y,z)=\xi \left \| P_{current}-P_{goal} \right \|^{2},
\end{equation}
where $U(x,y,z)$ is the overall potential, $U_{a}(x,y,z)$ is the attraction potential. The attraction potential is calculated as a function of current position $P_{current}$ and the goal position $P_{goal}$, where $\xi$ is a scaling factor. The obstacle repulsive potential $U_{r}(x,y,z)$ is only active when the drone is located within $d_\text{0}$ distance from the obstacle and is given by Eq. \ref{eq:5}:

\begin{equation}
 \label{eq:5}
 U_{r}(x,y,z) = \begin{cases}
 \frac{1}{2}\eta(\frac{1}{\rho(x,y,z)} - \frac{1}{d_\text{0}})^2 \quad &, \hspace{2mm}\rho \leq d_\text{0} \\
 0 \quad & ,\hspace{2mm}\rho > d_\text{0} \\
 
 \end{cases} ,
\end{equation}
where $\rho(x,y,z)$ is the chosen Euclidean distance function and $\eta$ is the scaling factor. The repulsive potential is inversely proportional to the distance between a drone and an obstacle. 

During the second mode of operation, one or more drones are blinded-flying without a camera feed. One of the remaining drones becomes the leader and shifts its forwarded-facing camera to a downward-facing position to be able to locate the blinded drones and to guide them to the landing pad on the moving platform. The leader drone then sends the location of the remaining drones and landing platform to the main computer. 
The leader drone is selected according to the cost function defined by  Eq. \ref{eq:6}.
\begin{equation}
 \label{eq:6}
f(p, q) =\sqrt{\sum_{i=1}^3(q_{i}-p_{i})^2}
\end{equation}
where $q_{i}$ is the position of the malfunctioning drone, $p_{i}$ is the current position of the drone with active camera.
The cost function is defined by the Euclidean distance from the current position of the drone to the position of the blind drone. APF is used to avoid collision between the follower drones while performing a landing. The swarm control algorithm is given by Algorithm \ref{alg:1}. 
\begin{algorithm}
\caption{Swarm Control Algorithm}\label{alg:1}
\begin{algorithmic}[1]
\State $Swarm \gets [Drone1, Drone2, Drone3]$ 
\While{$True$}
  \If{$all \, cameras$ are $working$}
       \State $Leader\, selection \gets False$
        \For{$Drone$ in $Swarm $ }
             \State $Drone.mode \gets mode_1$
        \EndFor 
    \Else
        \If{$Leader$ == $False$} 
             \State $Leader\, Drone$ = $min(cost(Swarm))$
             \State $Leader\, selection \gets True$
        \Else
             \For{$Drone$ in $Swarm $ }
                 \If{$Drone$ is $Leader$}
                    \State $Drone.mode \gets mode_2.leader$
                 \Else
                    \State $Drone.mode \gets mode_2.follower$
                 \EndIf
             \EndFor
         \EndIf
     \EndIf
\EndWhile
\end{algorithmic}
\end{algorithm}

\section{Experimental Setup}

Three experiments were carried out on landing a swarm of drones on a static platform, a platform moving along a line, and a platform moving along a complex trajectory (rectangle). In the first two experiments, two options for drone control were considered. In the first case, the drones found the landing site using their cameras. In the second case, the movement of a swarm of drones was considered in an emergency case of a breakdown of one or more cameras, then the landing site and the current positions of the drones were tracked from above by the camera of the leader drone. In the third experiment, the landing of the swarm in leader-follower formation was evaluated with platform moving along a complex trajectory. 

The size of the landing platform was 45$\times$57 cm. In the second and third experiments, the mobile platform moved at a speed of 0.5 m/s. In the leader-follower visual localization scenario the leader drone follows the mobile platform at a height of 2 meters with a lag of 0.5 meters in the horizontal plane. The speed of the drones was limited to 2.5 m/s, and the average speed along the complex trajectory was 0.7 m/s. The Vicon mocap system was used to obtain the trajectory of the mobile platform and the desired landing points for the drones. All experiments were aimed at calculating the RMSE of landing, each experiment for each case was carried out 5 times (overall 15 flight experiments were accomplished).

\begin{figure}[!h]
 \centering
 \includegraphics[width=1.0\linewidth]{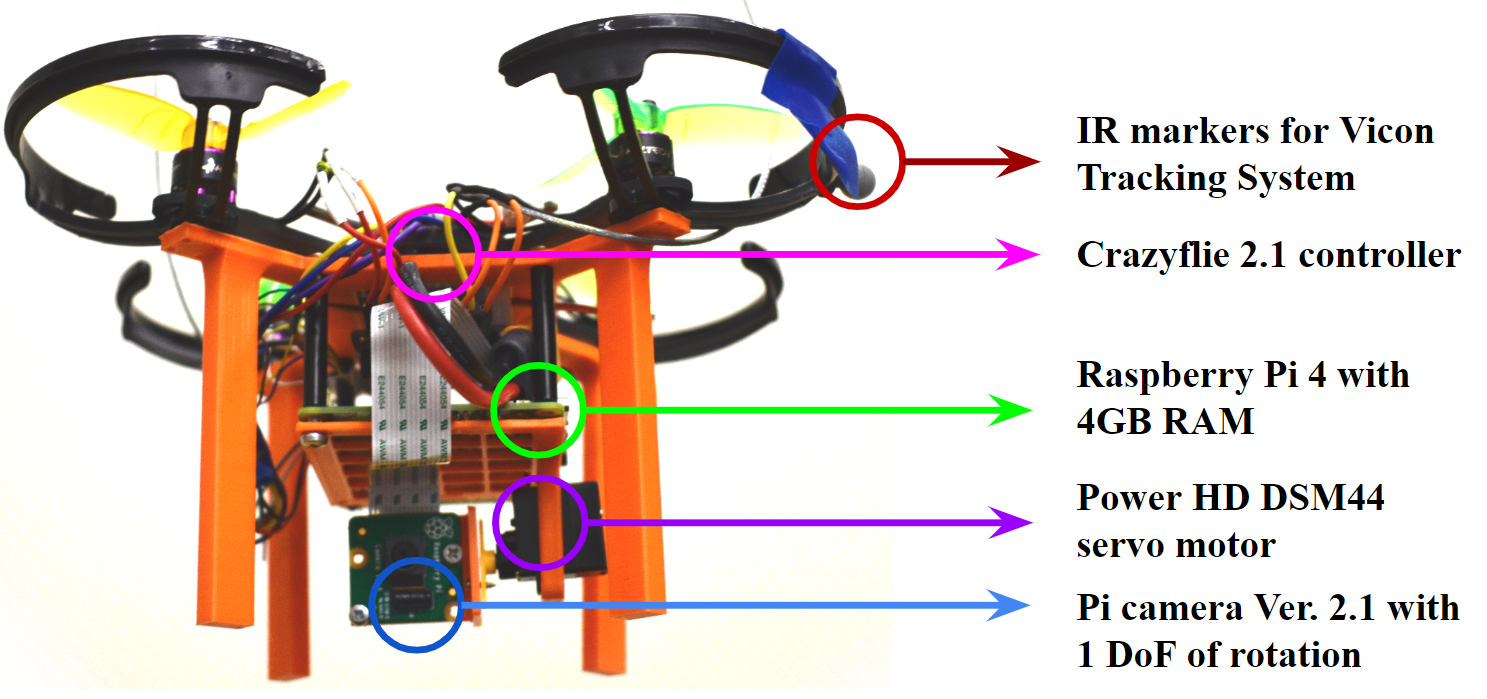} 
 \caption{Hardware components of the SwarmHawk. Pi camera is rotated by servo motor between forward-facing and downward-facing positions when operating in two modes.}
 \label{fig:drone_fpv}
\end{figure}

The swarm of drones used for the experiment is based on the Crazyflie platform (Fig. \ref{fig:drone_fpv}).

The leader drone weighs 262 grams, including a 450 mAh 3S battery. The follower drones weigh 32 grams including a 240 mAh 1S battery. The onboard controllers used for the leader drone are the Crazyflie 2.1 with a quad deck extension and Raspberry Pi 4 with Pi camera, which provides 640$\times$480 px RGB images at a rate of 30 fps to the target detection module, whereas the rest of the swarm carries on board only the Crazyflie 2.1 controller.

\subsection{Landing on Stationary Platform}
\subsubsection{Experimental procedure}
In this experiment, the mobile robot is set to a stationary position, and the swarm of drones is launched. In the first case, a swarm of three drones lands on the platform, receiving coordinates from image processing on their cameras. In the second case, the cameras of two drones fail, and one drone flies higher (by 2 m), and the camera located horizontally reads the coordinates of the drones in the swarm and the coordinates of the platform.

\subsubsection{Experimental results}
\begin{figure}[!h]
 \includegraphics[width=1.0\linewidth]{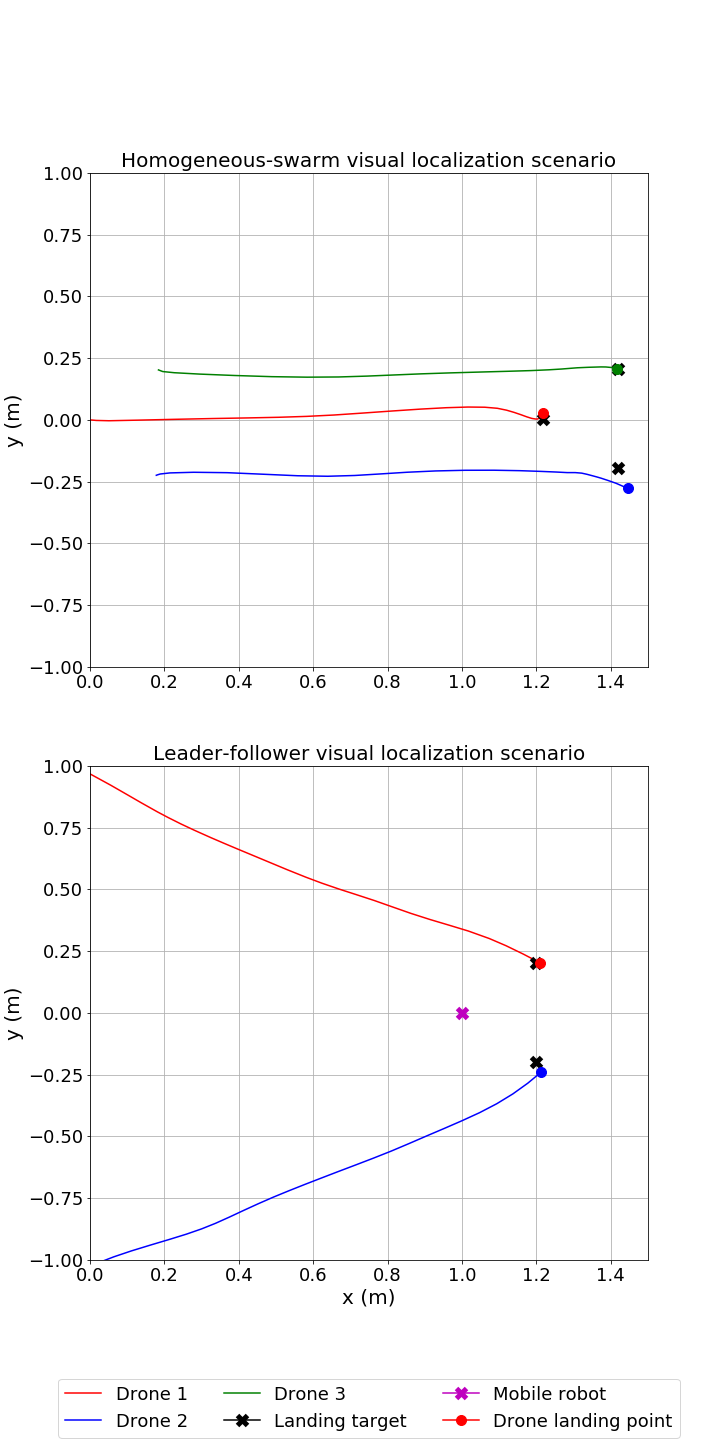}
 \caption{A top view of trajectories of SwarmHawk drones while landing on a stationary platform. Top picture (a) represents the homogeneous-swarm localization of three drones. Bottom picture (b) represents the leader-follower localization scenario, where a central drone navigated two followers. The purple cross is the position of the mobile platform center. Colored lines are the actual drone trajectories. Crosses represent the desired targets on a stationary platform.}
 \label{fig:stationary}
\end{figure}

The experimental results are presented in Fig. \ref{fig:stationary}.

The overall error of swarm formation landing was 4.2 cm in the case where each drone has a sight. The experimental results revealed that the drones landed within proximity to the target.

The better landing result was shown in the case where the leader drone is tracking the sightless drones. The average landing error in the leader-follower visual localization scenario is 1.9 cm, which is much more accurate than the previous scenario. The result could be explained by the fact that the cameras in the first scenario generates data with higher oscillation and noise level while moving. In the second case while landing on a stationary platform the leader drone is in a stable position, and the data from the camera is more accurate.

\subsection{Landing on a Moving Platform}
\subsubsection{Experimental procedure}
In this experiment, we tested two swarm landing scenarios on a moving platform. The landing platform was moving at a constant speed of 0.5 m/s in a straight line. As the moving platform moves, the swarm of drones follows the robot's trajectory from its initial positions without changing their landing formation by yawing until they land on the target location.


\subsubsection{Experimental results}
\begin{figure}[!h]
 \includegraphics[width=1.0\linewidth]{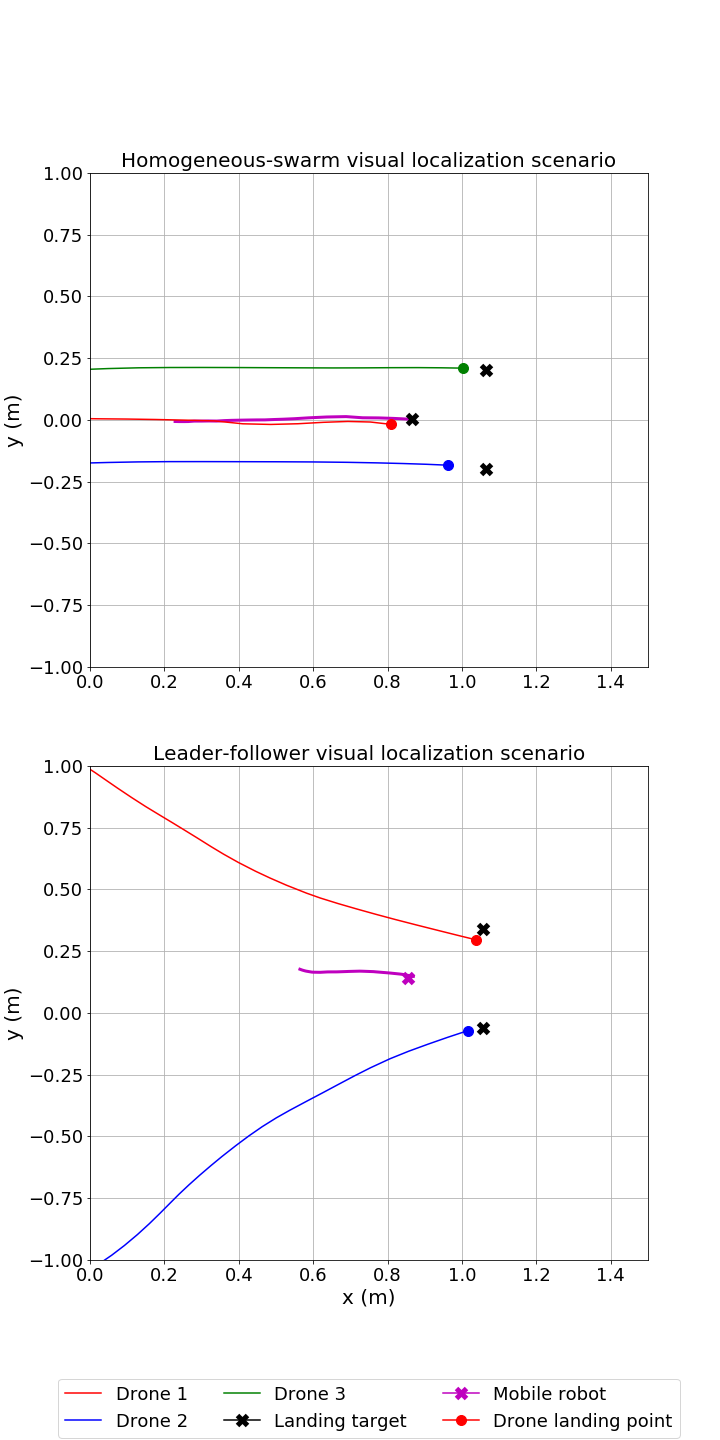}
 \caption{A top view of trajectories of SwarmHawk drones while landing on a moving platform. Top picture (a) represents the homogeneous-swarm localization of three drones. Bottom picture (b) represents the leader-follower localization scenario, where the central drone navigated two followers. Purple cross and line are the position and trajectory of the mobile platform center. Colored lines are actual drone trajectories. Crosses represent the landing targets on the moving platform.}
 \label{fig:line}
\end{figure}

The experimental results showed a minor landing error in the leader-follower visual localization scenario (Fig. \ref{fig:line}). In this condition, we can observe that as the drones identify the target tag, they start to move towards it and switches to the landing state as the threshold is achieved. The overall error for the swarm formation is 6.9 cm in the homogeneous-swarm visual localization scenario and 4.7 cm in the leader-follower visual localization scenario.

\subsection{Landing on a Platform Moving along a Complex Trajectory}
\subsubsection{Experimental procedure}

In the last experiment, the drones landed on a platform moving along a complex trajectory with a constant speed of 0.5 m/s in the leader-follower visual localization scenario, since it showed less landing error in the first two experiments. When the platform moved along the rectangle, the drones followed the tag and changed the yaw angle according to the tag's rotation in the frame of the leader drone's camera. At the end of the trajectory, the drones land on the mobile platform.

\subsubsection{Experimental Results}
\begin{figure}[!h]
 \includegraphics[width=1.0\linewidth]{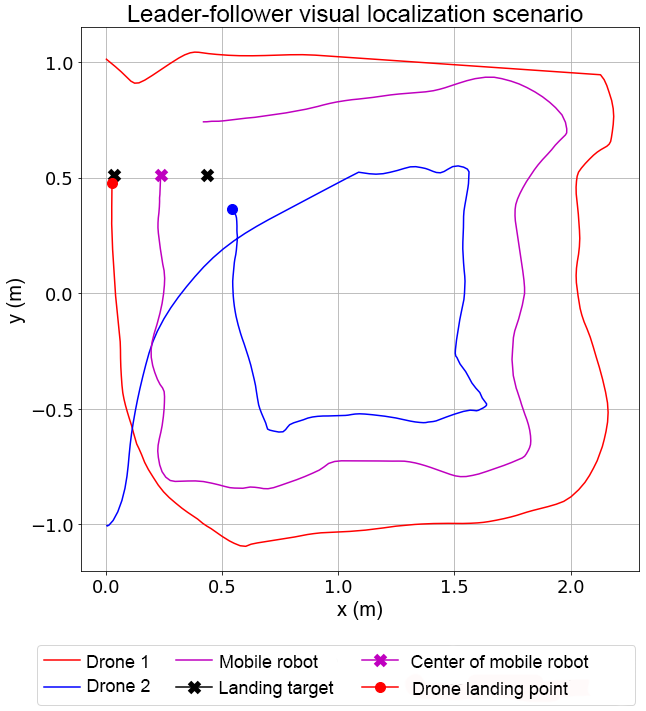}
 \caption{A top view of trajectories of SwarmHawk drones while landing on a mobile platform, moving by a rectangular trajectory. The picture represents the leader-follower localization scenario, where the central drone navigated two followers. Purple cross and line are the position and trajectory of the mobile platform center. Colored lines are actual drone trajectories. Crosses represent the landing targets on the moving platform.}
 \label{fig:rect}
\end{figure}

\begin{figure}[!h]
 \includegraphics[width=1.0\linewidth]{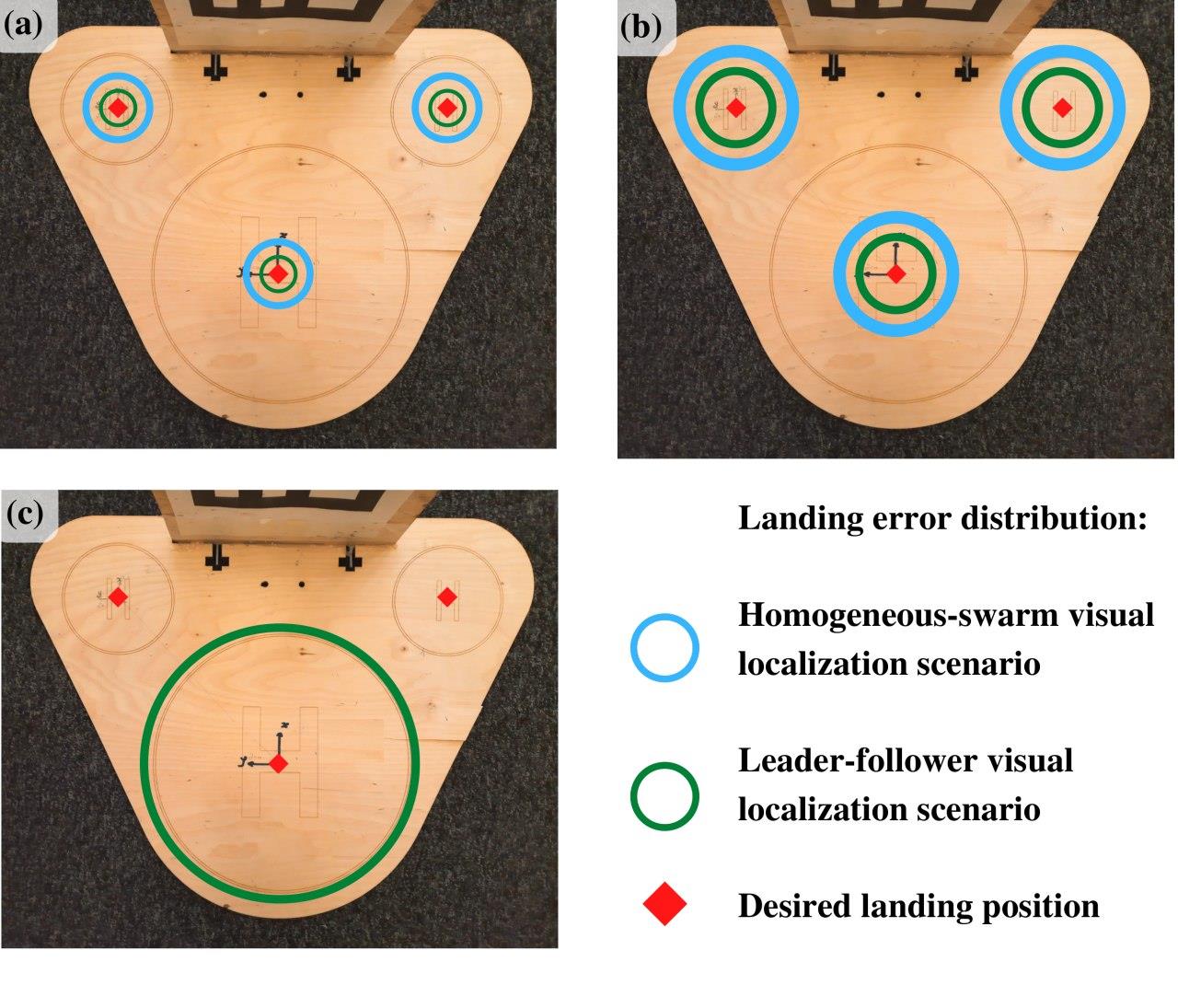}
 \caption{Drone landing boundaries for: a) Landing on a stationary platform. b) Landing on a moving platform. c) Landing on a platform moving along a complex trajectory.}
 \label{fig:all_error}
\end{figure}

Since the drones followed the mobile robot along a complex trajectory for a long time and rotated along the z-axis to maintain their formation (Fig. \ref{fig:rect}), in this experiment the maximum error was higher than in the case of the stationary platform. However, the average positional error did not exceed 19.4 cm, which was considered a sufficient accurate by not exceeding the dimensions of the landing platform (Fig. \ref{fig:all_error}). In the future, it is required to improve the swarm behavior at the last stage of landing for high-speed vehicles.

\section{Conclusions and Future Work}

We have developed a novel drone swarm docking system SwarmHawk to allow landing of swarm of drones on a moving platform when one of them lost the position signal. The key feature of the technology is its ability to switch in real-time between the homogeneous-swarm mode and the leader-follower mode, depending on the localization state of the entire system. The AFP method was applied to the follower drones to reach the vision tag while avoiding internal swarm collision. The CV-based system was developed and tested to detect the visual marks on the mobile platform and swarm agents and calculate their distance from the leader drone. The experimental results revealed a low landing error in the case of static (of 1.9 cm) and moving (of 4.7 cm) platforms for both states of the swarm. Moreover, the increase of error in the homogeneous-swarm visual localization scenario was not critical. Moreover, the drones showed a good landing on a platform moving along a complex trajectory (average error of 19.4 cm) in the leader-follower visual localization scenario.

The proposed technology can be applied in the remote exploration and mapping scenarios in GPS-denied areas and under harsh conditions, which can affect ground station performance and cause the fault of several swarm agents.

In future work, we plan to address different approaches to the task dispatching in the swarm to achieve an effective way of the hierarchy control and leader drone assignment. Moreover, the algorithms for the proactive guidance with vehicle direction estimation during drone landing will be developed and evaluated to decrease the error during the landing on platforms moving along a complex trajectory. The effects of the external disturbances (wind, turbulence, etc.) on our system will also be experimentally evaluated. One of the interesting research will be conducted to achieve landing of drones on another UAV in flight to perform for example battery recharging. 

\section*{Acknowledgment}
The reported study was funded by RFBR and CNRS, project number 21-58-15006.

\bibliographystyle{IEEEtran}
\bibliography{sample-base}
\end{document}